# From Patterson Maps to Atomic Coordinates: Training a Deep Neural Network to Solve the Phase Problem for a Simplified Case

David Hurwitz[*,~]

**Abstract**

This work demonstrates that, for a simple case of 10 randomly positioned atoms, a neural network can be trained to infer atomic coordinates from Patterson maps. The network was trained entirely on synthetic data. For the training set, the network outputs were 3D maps of randomly positioned atoms. From each output map, a Patterson map was generated and used as input to the network. The network generalized to cases not in the test set, inferring atom positions from Patterson maps.

A key finding in this work is that the Patterson maps presented to the network input during training must uniquely describe the atomic coordinates they're paired with on the network output or the network will not train and it will not generalize. The network cannot train on conflicting data. Avoiding conflicts is handled in 3 ways: 1) Patterson maps are invariant to translation. To remove this degree of freedom, output maps are centered on the average of their atom positions. 2) Patterson maps are invariant to centrosymmetric inversion. This conflict is removed by presenting the network output with both the atoms used to make the Patterson Map and their centrosymmetry-related counterparts simultaneously. 3) The Patterson map does not uniquely describe a set of coordinates because the origin for each vector in the Patterson map is ambiguous. By adding empty space around the atoms in the output map, this ambiguity is removed. Forcing output atoms to be closer than half the output box edge dimension means the origin of each peak in the Patterson map must be the origin to which it is closest.

# 1. Introduction

## 1.1 The Phase Problem

X-ray crystallography has been an essential tool in protein and nucleic acid structure determination since the 1950s[1,2]. Of the 158,367 structures in the Protein Data Bank in December 2019, 89% were solved using X-ray crystallography[3].

A crystal in the path of an X-ray beam produces an X-ray diffraction pattern. The diffraction pattern is the Fourier transform of the electron density in the crystal's repeating unit. However, the data that can be collected is incomplete. The magnitude of each spot can be measured, but the phases cannot be recovered. The loss of phase information during an X-ray crystallographic experiment is known as "the phase problem".

Were the phases recoverable, it would be a simple matter to reconstruct the molecular structure of the crystal's repeating unit by computing the inverse Fourier transform of the diffraction data. Absent phases, other strategies, such as molecular replacement and multiple isomorphous replacement, are used to determine macromolecular structure from diffraction magnitudes.

In some cases, it is possible to compute molecular structure directly from X-ray data. This is referred to as "direct methods" in X-ray crystallography. Direct methods make use of the fact that, when electron density maps are composed of atoms, and density is always positive, the diffracted phases are not random. Rather, phases of certain reflection triplets have a high probability of summing to zero[4]. Solving a system of such probability equations can determine the unknown phases[5]. In practice, this has only been possible for molecules of up to a few hundred atoms that diffract to high resolution.

This work is a first attempt at using a neural network as an alternative to direct methods. The idea is to see if a neural network can substitute for solving the probability equations of direct methods. If this is possible for small molecules at high resolution, it would then be interesting to see if a neural network can be scaled up to larger molecules, or trained for lower resolutions, than direct methods currently allow.

## 1.2 The Patterson Function

One useful tool to help interpret a diffraction pattern is the Patterson Map[6]. The Patterson Map is the inverse Fourier Transform of the *square* of the magnitudes of the diffraction data. Whereas a Fourier Transform of magnitudes and phases gives molecular structure, the Fourier Transform of the square of the magnitudes gives a map of vectors between each pair of atoms in the structure. This may be a more intuitive starting point than diffraction data.

To say that the Patterson Map is the inverse Fourier transform of the diffraction magnitudes squared is equivalent to saying that the Patterson Map is the original electron density map convolved with its inverse. This is a restatement of The Convolution Theorem: multiplication in Fourier space is equivalent to convolution in real space. Squaring the diffraction magnitudes in Fourier space, then doing an inverse Fourier transform on the product, is equivalent to doing a convolution of the electron density map with itself.

So, the problem of going from a Patterson Map to atomic coordinates is a deconvolution. In this light, we can see that the problem of going from Patterson Maps to atomic coordinates is, in some ways, comparable to other deconvolution problems that have already been examined with neural networks, such as image sharpening, and image "super-resolution"[7,8].

## 1.3 Convolutional Neural Networks

Neural networks are a shift from traditional rules-based programming. They learn to solve a problem by example, as opposed to solving a problem by following a comprehensive set of logical operations. A neural network is trained to do a task by showing the network, typically, many thousands of training examples, then seeing if the network can generalize for cases not in the training set.

A neural network is composed of highly interconnected nodes, arranged in layers, with weights that multiply the connection strengths between nodes. The connections are summed in the nodes, and a non-linearity is applied to each node's output. During training, the weights must be adjusted so that computed neural network outputs match known outputs from the training set. An important advancement in modern neural network methods was the development of learning by back-propagation of errors, for training the networks.[9]

When neural network layers are "fully connected", the number of weights in the network can be prohibitively large. In a fully connected layer, each node in the layer is connected to each node in the previous layer. For networks where the input to the network is, say, a high-quality image, and that rough image size is propagated through the network, the number of weights in a fully connected layer is the square of the image size, which is usually unworkable.

For this reason, the development of convolutional neural networks was critically important to developing effective neural networks for imaging applications[10]. In convolutional neural networks, convolution operations are performed on small portions of a layer's nodes at a time. Convolution "kernels" are typically 3 to 7 pixels per dimension. Many kernels can be used at each layer, but, still, the number of weights in the network can be quite small since the number of weights is proportional to the size of the kernel, not the square of the number of pixels in an image. The kernels are applied across an entire image, one patch at a time. So, the kernel weights are effectively shared across the image. Convolutional neural networks have enabled neural networks to be built with a relatively small number of weights at each layer, and a large number of layers. This type of network architecture was another important advancement in the deep-learning revolution.

At the 2012 ImageNet Large Scale Visual Recognition Challenge, convolutional networks demonstrated impressive progress at classifying images. The error rate in this competition dropped from 26% in 2011 to 16% in 2012 due to the deep (8 layer) convolutional neural network architecture known as AlexNet introduced that year[11]. In the following few years, error rates in this competition declined to just a few percent, matching human performance. From 2012 on, all winners used improved convolutional neural network architectures. To some, the 2012 ImageNet competition marked the start of the AI boom[12].

One thought behind this work is that by presenting the phase problem to the neural network as images, with Patterson maps on the input, and simulated electron density maps on the output, this work can benefit from some of the improvements in image processing that have been achieved in the last few years, particularly the use of convolutional neural networks.

## 2. Methods

### 2.1 Output Data Representation

The neural network output images are simplified versions of electron density. For the training set, 10 random "atoms" are placed in a box such that the atoms are non-overlapping. The atomic radius for each atom is 1-pixel length. The value assigned to each pixel is the fraction of the pixel that is occupied by the sphere. The work is done in three dimensions, but a two-dimensional example is shown in Figure 1.

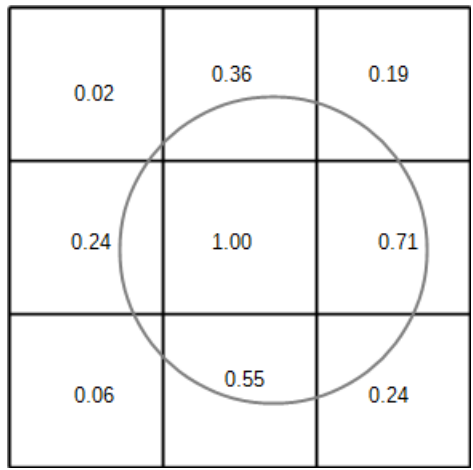

Fig. 1. This figure shows a two-dimensional example of the neural network output. In this work, spheres are placed on voxels, but this figure illustrates the method in 2D. Each voxel's value is the fraction of the voxel that is occupied by a randomly positioned sphere of radius 1-pixel length. Values in this figure are not precise.

The neural network described here was trained with input and output images of size [40 x 40 x 40] pixels³. Roughly 20 pixels were occupied for each atom. So, the output matrices are quite sparse. For 10 atoms, about (20 · 10) / (40 · 40 · 40) = 0.3% of pixels have non-zero values. However, due to considerations discussed in Methods 2.4.2 and 2.4.3, 20 atoms are used and are confined to a smaller space, giving a less sparse (20 · 20) / (14 · 14 · 14) = 15% of pixels in the smaller box that have non-zero values.

## 2.2 Input Data Representation

The neural network inputs are the Patterson maps of the outputs. Their dimensions are the same as the output maps. The Patterson map is the Fourier transform of the diffraction magnitudes squared. This is equivalent to making the Patterson map by convolving the electron density map with its inverse as discussed in the Introduction 1.2. In this work, the Patterson map is computed by taking the Fourier Transform of the output density map, multiplying the Fourier Transform by its complex conjugate, and then taking the inverse Fourier Transform of the product. This equivalence is show below.

(1) $P(u) = \rho(r) \otimes \rho(-r)$
(2) $F(P(u)) = F(\rho(r)) \cdot F(\rho(-r))$
(3) $F(P(u)) = F_h \cdot F_h^*$
(4) $F(P(u)) = |F_h|^2$
(5) $P(u) = F^{-1}(|F_h|^2)$

Equation (5), the Patterson function, states that the Patterson map is the inverse Fourier Transform of the square of the magnitudes of the diffraction data. $F_h$ are crystallographic reflections. Working backwards, the Patterson Function can also be computed by convolving electron density with its inverse (eq. 1). One can take the Fourier Transform of each side and invoke the Convolution Theorem to give eq (2). Since the Fourier Transform is conjugate symmetric (f(−x)=f*(x)), we get that the Fourier Transform of the Patterson map is the diffraction data times its complex conjugate (eq. 3), which is equivalent to squaring the diffraction magnitudes (eq. 4).

## 2.3 The Network Architecture

I used the Keras platform[13] for building and training the neural network in this project. An Nvidia GTX 1080 Ti GPU was used to accelerate training and inference. The neural network architecture for this work is straightforward. A code snippet is shown:

```
input  =   Input(shape=data.InData.shape)
L01a   =   Conv3D(20, 5, activation='relu', padding='same', kernel_initializer='he_normal')   (input)
L01b   =   Conv3D(20, 5, activation='relu', padding='same', kernel_initializer='he_normal')   (L01a)
D02    =   MaxPooling3D(pool_size=(2,2,2))                                                     (L01b)
L08a   =   Conv3D(20, 7, activation='relu', padding='same', kernel_initializer='he_normal')   (D02)
L08b   =   Conv3D(20, 7, activation='relu', padding='same', kernel_initializer='he_normal')   (L08a)
L08c   =   Conv3D(20, 7, activation='relu', padding='same', kernel_initializer='he_normal')   (L08b)
L08d   =   Conv3D(20, 7, activation='relu', padding='same', kernel_initializer='he_normal')   (L08c)
L08e   =   Conv3D(20, 7, activation='relu', padding='same', kernel_initializer='he_normal')   (L08d)
L08f   =   Conv3D(20, 7, activation='relu', padding='same', kernel_initializer='he_normal')   (L08e)
L08g   =   Conv3D(20, 7, activation='relu', padding='same', kernel_initializer='he_normal')   (L08f)
L08h   =   Conv3D(20, 7, activation='relu', padding='same', kernel_initializer='he_normal')   (L08g)
U03    =   UpSampling3D(size=(2,2,2))                                                          (L08h)
L15a   =   Conv3D(20, 5, activation='relu', padding='same', kernel_initializer='he_normal')   (U03)
output =   Conv3D( 1, 5, activation='tanh', padding='same', kernel_initializer='he_normal')   (L15a)
model  =   Model(inputs=[input], outputs=[output])
print(model.summary(line_length=150))
model.compile(optimizer=Adam(lr=LR), loss=customLoss)
```

There are 12 3D convolutional layers in this architecture that use kernels of either [5 x 5 x 5] or [7 x 7 x 7] pixels[3]. Each layer, aside from the output layer, has a set of 20 output kernels. The input image size is [40 x 40 x 40]. One max-pooling layer shrinks this down by a factor of 8 to [20 x 20 x 20], and one up-sampling layer expands the dimensions back to their original size.

The number of weights in the network is:
$5 \cdot 5 \cdot 5 \cdot 20 \cdot 1 = 2,500$ for 2 layers (the first and last layers)
$5 \cdot 5 \cdot 5 \cdot 20 \cdot 20 = 50,000$ for 2 layers
$7 \cdot 7 \cdot 7 \cdot 20 \cdot 20 = 137,200$ for 8 layers
$(2,500 \cdot 2) + (50,000 \cdot 2) + (137,200 \cdot 8) + (20 \cdot 11 + 1)$ (the bias terms) = 1,202,821 weights.

The number of weights in the down-sampled layers is about $(7 \cdot 7 \cdot 7) / (5 \cdot 5 \cdot 5) = 2.7$ times the number of weights in the higher resolution layers. This is in rough keeping with the pattern of doubling the number of channels when down-sampling with a fixed kernel size[14].

The receptive field for this network is: $(2 \cdot 4) + (2 \cdot 3 \cdot 8) = 56$. It was necessary to have a receptive field at least the size of the input image, as the output image is known to be an auto-correlation of the input image. A receptive field at least the size of the input image is needed so that each network output pixel "feels" each network input pixel.

A schematic for the network architecture used in this work is show in figure 2.

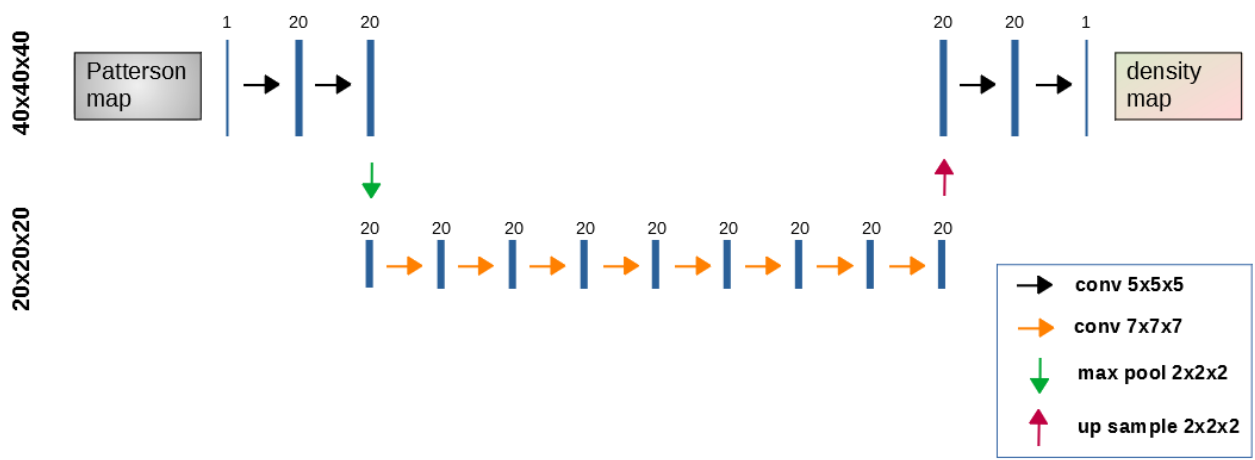

Fig. 2. This is a schematic representation of the neural network architecture used in this work. The input and output images are [40 x 40 x 40] pixels[3]. The 1 channel output is an electron density map, and the 1 channel input is a Patterson map. At high resolution, 4 3D convolutions are applied. In between, the network image is down-sampled to [20 x 20 x 20] pixels[3] where 8 3D convolutions are applied before the image is up-sampled back to its original dimensions. Each layer, aside from the output layer, uses a set of 20 kernels.

Network performance over small-scale trial runs guided hyperparameter optimization, including optimization of kernel sizes, activation functions, number of layers, batch size, and so on.

No batch normalization layers were used in this network. I found that, because of very sparse output data (see Introduction 1.3), the batch normalization layers led to problems in training, even when using them only in the early layers of the network. Without batch normalization layers, I decided to keep the number of layers relatively low. This was a large part of the reason for adding the down-sampled layers. In addition, the down-sampled layers have the added advantage of speeding up training and increasing the network receptive field. On the other hand, their use probably resulted in decreased accuracy measuring atom positions.

## 2.4 Important Considerations When Training the Network

I found there are 3 important considerations to take into account when training the network, or the network will not train and it will not generalize. These considerations all boil down to the same main point: a neural network cannot train on conflicting data. For example, an image classifier cannot train if images in the training set are initially labeled one way, and then the labels are changed. You cannot tell a network: here is an image of a cat, then show the network the same image and tell the network the image is now a dog. Likewise, when training a neural network to infer atomic coordinates from Patterson maps, you cannot show the network a Patterson map on the input, with one arrangement of atoms on the output, then show the network the same Patterson map, and on the output show it a different atomic arrangement. These 3 considerations are discussed in Methods 2.4.1 – 2.4.4.

## 2.4.1 Removing Translational Degrees of Freedom

The Patterson map is translation invariant. A set of atomic coordinates has the same Patterson map regardless of its translation. For a neural network to train, coordinates on the output of the network must be translationally fixed. If not, the network could be presented with multiple training cases that have the same Patterson map on the input, but different translations of the atoms on the output. In this work, removing translational freedom was accomplished by translating the atoms presented to the neural network output so the average atom position is at the center of the output map.

## 2.4.2 Removing Centrosymmetry Ambiguity

The Patterson map is invariant to centrosymmetric inversion. This means a set of atoms has the same Patterson map as its centrosymmetry-related atoms. This is problematic for the network to train and generalize. For this work, the problem is solved by training on **both** sets of atoms **simultaneously**. Without this technique, the Patterson map is not unique as it describes 2 set of atoms. By training on both sets of atoms simultaneously, this ambiguity is removed. However, during the inference stage, the network finds twice as many atom positions as desired, and the 2 sets of atoms must be untangled. See Methods 2.4.4 for the method used in this work to separate 2 sets of atoms.

Whereas the initial 10 atoms were chosen to be non-overlapping, this condition was dropped when doubling the number of atoms in the output map. Atom clashes are allowed. To accommodate the possibility of clashes, electron density occupancy maps (see Methods 2.1) were scaled so the maximum is 0.5 rather than 1.0, allowing for combined density of 1.0 for 2 clashing atoms. An example of an output map is shown in figure 3.

The method works as follows: A set of 10 random atom positions are generated. The electron density for this set is calculated (density A). The Patterson map of density A is also calculated. This Patterson map is the neural network input for one training example. 10 atoms related by centrosymmetry to the original 10 atoms are calculated. The electron density for these 10 atoms is also calculated (density B). Density A and density B are summed, and this sum is used as the neural network output. In this way, 2 density maps that both produce the input Patterson map are presented to the neural network output as one training example so the network trains on both cases simultaneously.

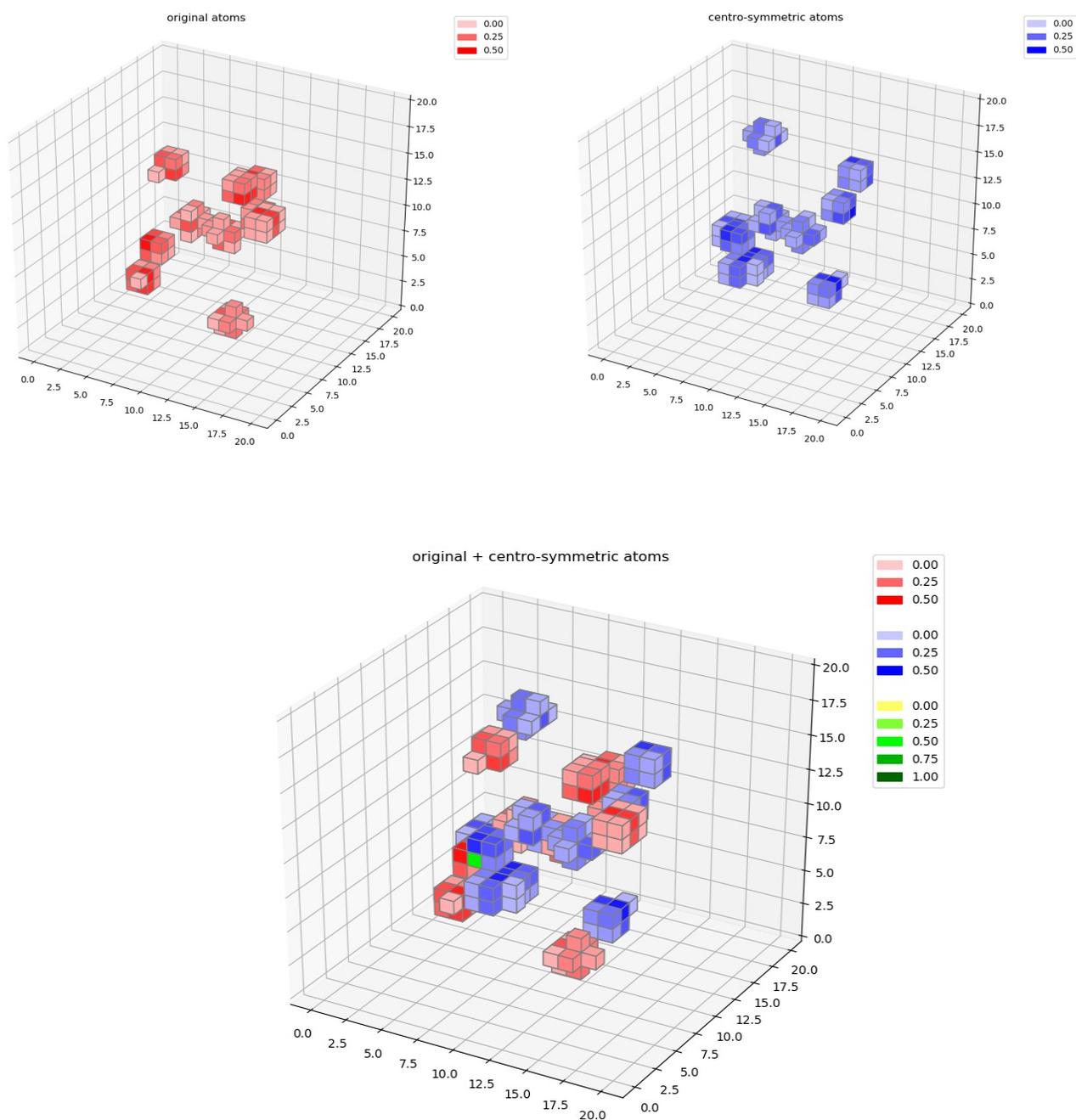

Fig. 3. This figure shows an example of an output density map used for training. The red density (upper left) shows the original 10 atoms. The Patterson map presented to the neural network input is calculated from this density map. The blue density (upper right) shows the atoms related by centrosymmetry. The 3rd image shows the combined density map for 20 atoms used for training. Clashes between original and symmetry-related atoms are allowed, as shown in green in the 3rd image.

### 2.4.3 Removing Ambiguity about Vector Origins

With respect to the third important consideration for training this neural network, I found that it is necessary to add empty space around the atoms for the network to train and generalize. It must be that without this change, the Patterson map on the input does not uniquely determine the atomic coordinates on the output. A simple example of the non-uniqueness of a Patterson map is shown in figure 4.

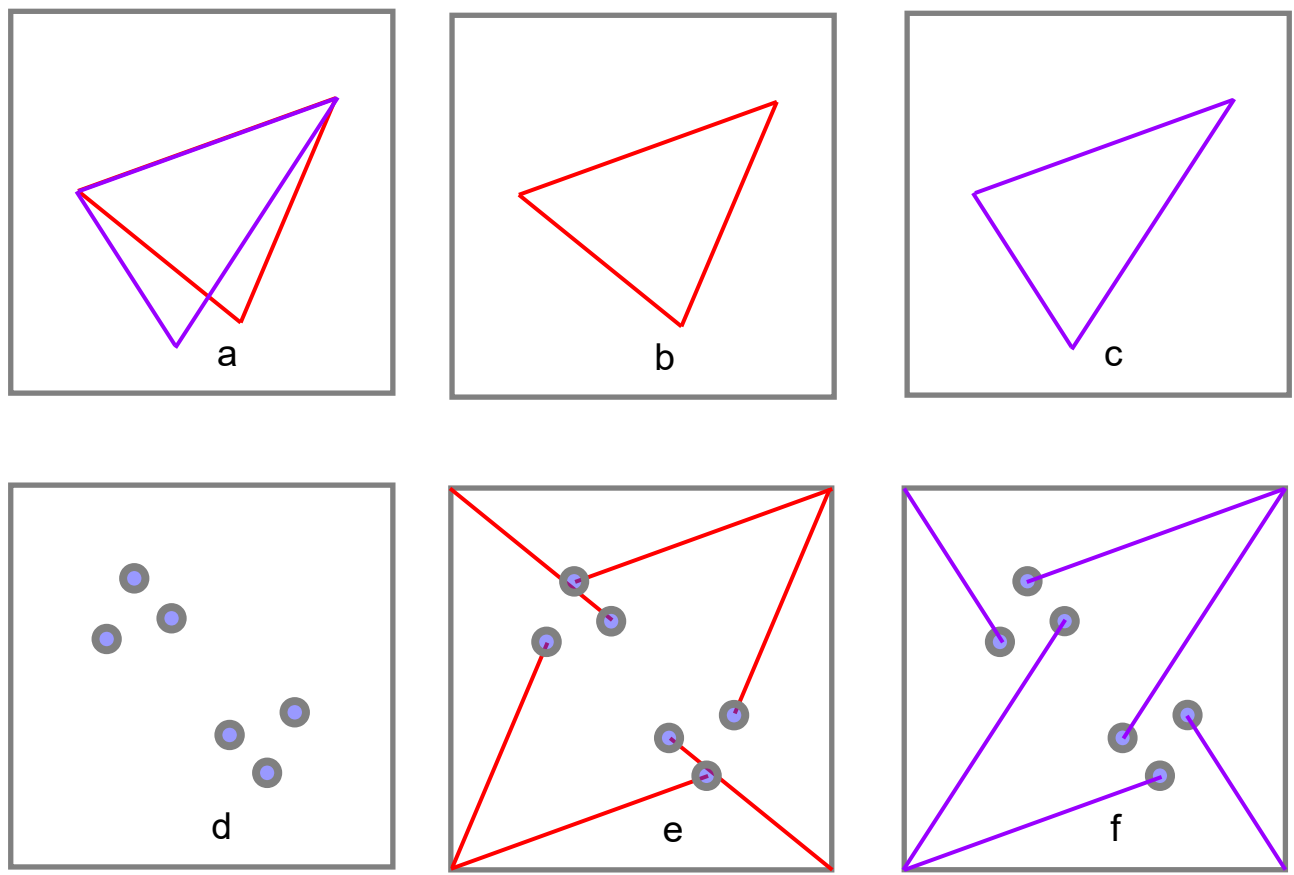

Fig. 4. Two sets of 3 atoms (b & c) clearly have different spacial arrangements (a). Yet, when the vectors between the atoms are drawn (e & f), the end-points are identical (d), demonstrating that more than one arrangement of atoms produce the same Patterson map.

By adding empty space around the atoms, the maximum distance between atoms can be limited to half the edge-distance of the output box. This removes ambiguity about the vector origin for each peak in the Patterson map. In the example shown in figure 5, the outer box represents the volume that is presented to the neural network output. The atoms are confined to an inner box whose dimension is chosen so that the maximum distance between atoms is half the edge-distance of the output box. In this way, when a peak is seen in the Patterson map, the vector for that peak must originate at the corner to which it is closest, as the distance from the peak to the other corners is greater than the maximal distance between atoms, as shown in figure 6. In this way, the inner box constraint eliminates all but one possible vector origin for each Patterson peak.

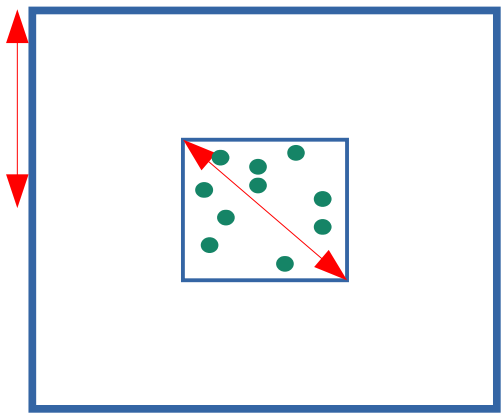

Fig. 5. In this work, the randomly positioned atoms presented to the neural network output were confined to a box within a box. The outer box shown here represents the full volume that is presented to the neural network output, and the inner box is chosen so that the maximum distance between any pair of atoms is half the edge-distance of the outer box. The 2 red distance measures shown here have the same length of half the outer edge which is also $\sqrt{3}$ times the inner box dimension.

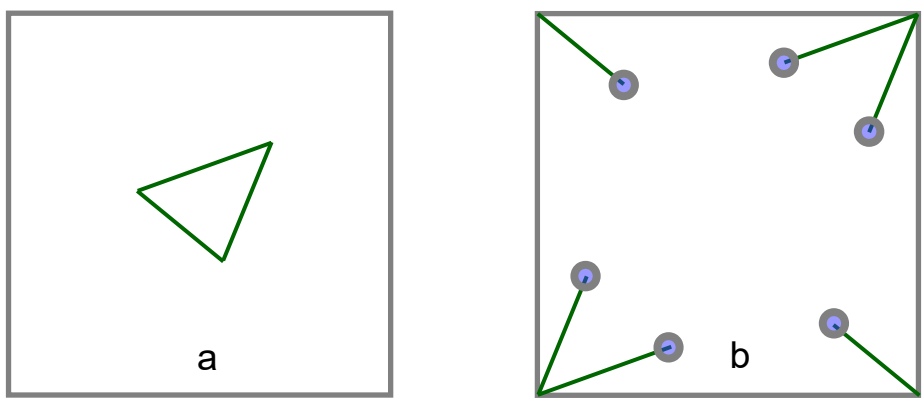

Fig. 6. When all distances between atoms are less than half the edge-distance of the output box (a), the interpretation of the Patterson vectors is unambiguous. The peaks must represent vectors originating from the corners they are closest to (b), as the distances to the other corners exceed the known inter-atomic distances.

In this work, the size of the output density is [40 x 40 x 40] pixels$^3$. For all training cases, the centers of 10 random atoms and their centrosymmetry-related atoms were confined to a smaller box of [12 x 12 x 12] pixels$^3$. Therefore, the maximal distance between atoms for the training cases was $12\sqrt{3}$ = 20.8 pixels, or roughly half the edge-distance of the output box.

By training the neural network with output density maps that have empty space surrounding the electron density, the network was able to train and generalize. This constraint reduces the number of interpretations of the Patterson map on the neural network's input.

### 2.4.4 Separating 2 Sets of Centrosymmetric Atoms

Training on 2 sets of atoms poses a problem for the inference stage. Since the network solves for both sets of atoms, these atoms need to be split into 2 sets: the original atoms plus the symmetry-related atoms. Fortunately, the Patterson map can be used as a guide for doing this. The algorithm used in this work to accomplish the separation was simple and effective:

- Locate 20 or so peaks in an output density map. Estimate atom positions from these peaks by averaging a peak pixel position and the positions of its neighbors, weighted by the density of each pixel.
- Double the number of atoms by adding centrosymmetic positions to the set of 20 positions.
- Choose 10 of the positions at random
- From the 10 positions, make a test density map
- Calculate a test Patterson map from the test density map
- Calculate a similarity score between the test Patterson map and the true Patterson map
- Repeat many times:
    - Swap one position in the set of 10 with a position not in the set of 10 but in the set of 40
    - Modify the test density map to reflect this change
    - Calculate a new test Patterson map for the new test density map
    - Calculate a similarity score between the new Patterson map and the true Patterson map
    - If the similarity score improved keep the swap
    - Otherwise, revert the swap and revert the test density map modification

I added simulated annealing to the above, though later tests indicated this may not have been necessary. In practice, this worked well and was effective at splitting the atoms into 2 sets.

# 3. Results

## 3.1 Training the Network

The neural network in this work was trained for more than 4000 epochs, cumulatively, over 26 "runs". In this context, a run means training the neural network without interruption with a set of hyperparameters. Hyperparameters were adjusted each run. Typically, the training batch size for a run was 3000 and the validation batch size was 100. One epoch of training took about 2.5 minutes. Accounting for the time to make the density and Patterson maps, this network trained for a total of about (4 mins/epoch x 4400 epochs) ≈ 300 hours, or about 12.5 days. The validation batch was constructed at the start, and used for the entire run, while the training batch was constructed at the start, then remade every few epochs. As a result of this training schedule, the training loss exhibited a zigzag character. For clarity, rather than display losses after each epoch, the high frequency zigzags of the training loss were removed by displaying losses just once for each new set of training data.

Figure 7 shows the training and validation loss over all training runs. The loss was the mean squared error between the known and inferred density maps. This plot indicates that the neural network generalized to cases not in the training set. First, the validation loss decline was in tandem with the training loss. Secondly, new training cases were made every few epochs, and the starting loss for these cases declined in parallel with the validation loss. In other words, it was evident just by swapping in new training cases that the network was generalizing.

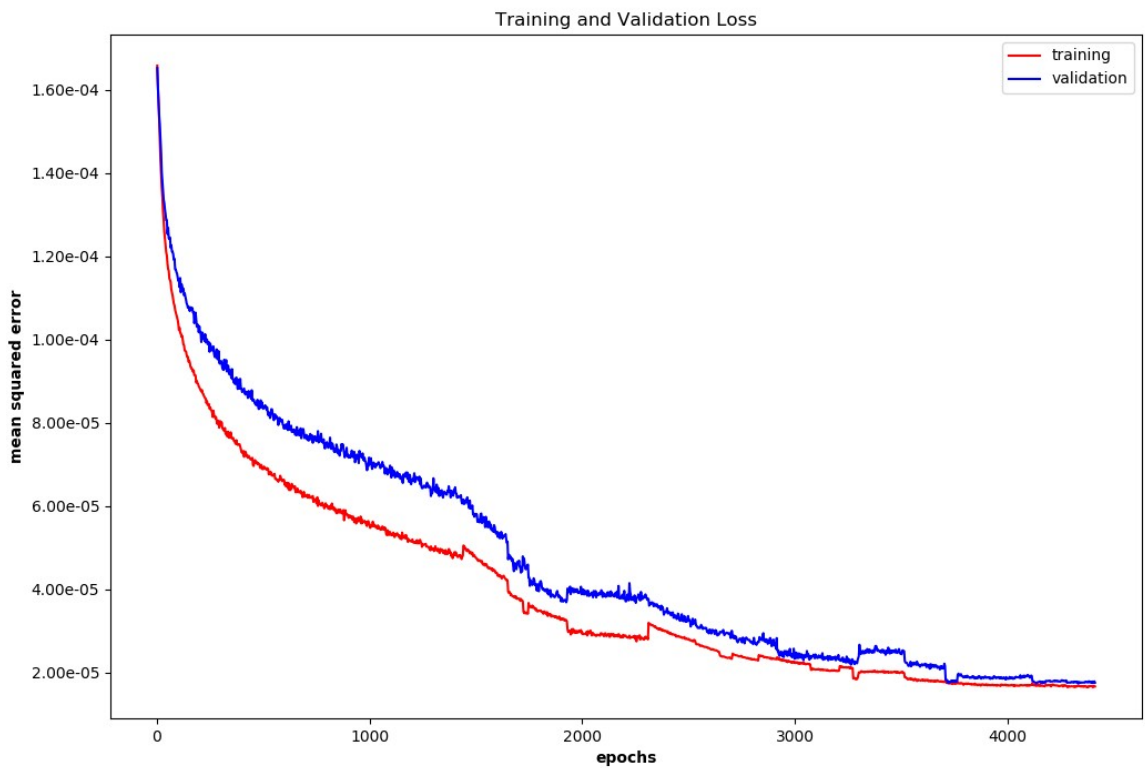

Fig. 7. The neural network in this work was trained for about 4000 epochs over 26 runs. Hyperparameters were adjusted each run, accounting for the erratic appearance of the combined plot. Typically, the training batch size was 3000 and was remade every 3 epochs. The training loss is displayed in red and the validation loss in blue. To remove high-frequency zigzags from this plot, only one training loss and one validation loss are displayed for each new set of training cases.

## 3.2 Trying to Train the Network when the Patterson Map is not Unique

I performed experiments to see what happens when the restrictions described in Methods 2.4.2 and 2.4.3 are relaxed. The network used throughout this work, and described in Methods 2.3, was trained from scratch 6 times for these 2 tests. Only the training data was changed, not the network architecture. Each time, the network was initialized to random values and trained for 100 epochs.

### 3.2.1 Training Without Centrosymmetric Atoms on the Output

For one test, one small change was made: rather than train on both the 10 original atoms and their 10 centrosymmetry-related atoms simultaneously, the network was trained on just original atoms. This test was done twice. In one case, the network was trained on 10 random atoms, and in another case, the network was trained on 20 random atoms. When the network was trained on 10 atoms, the measured loss was multiplied by 2, so that the loss could be compared to the loss when 20 atoms were in the network output. These tests, removing centrosymmetry atoms from the neural network output, were performed to test the claim that the network will not train or generalize unless the network trains simultaneously on both original atoms and their centrosymmetric counterparts. Results for this test are shown in figure 8, demonstrating that the network trains and generalizes much more effectively when training on both original atoms and centrosymmetry-related atoms simultaneously.

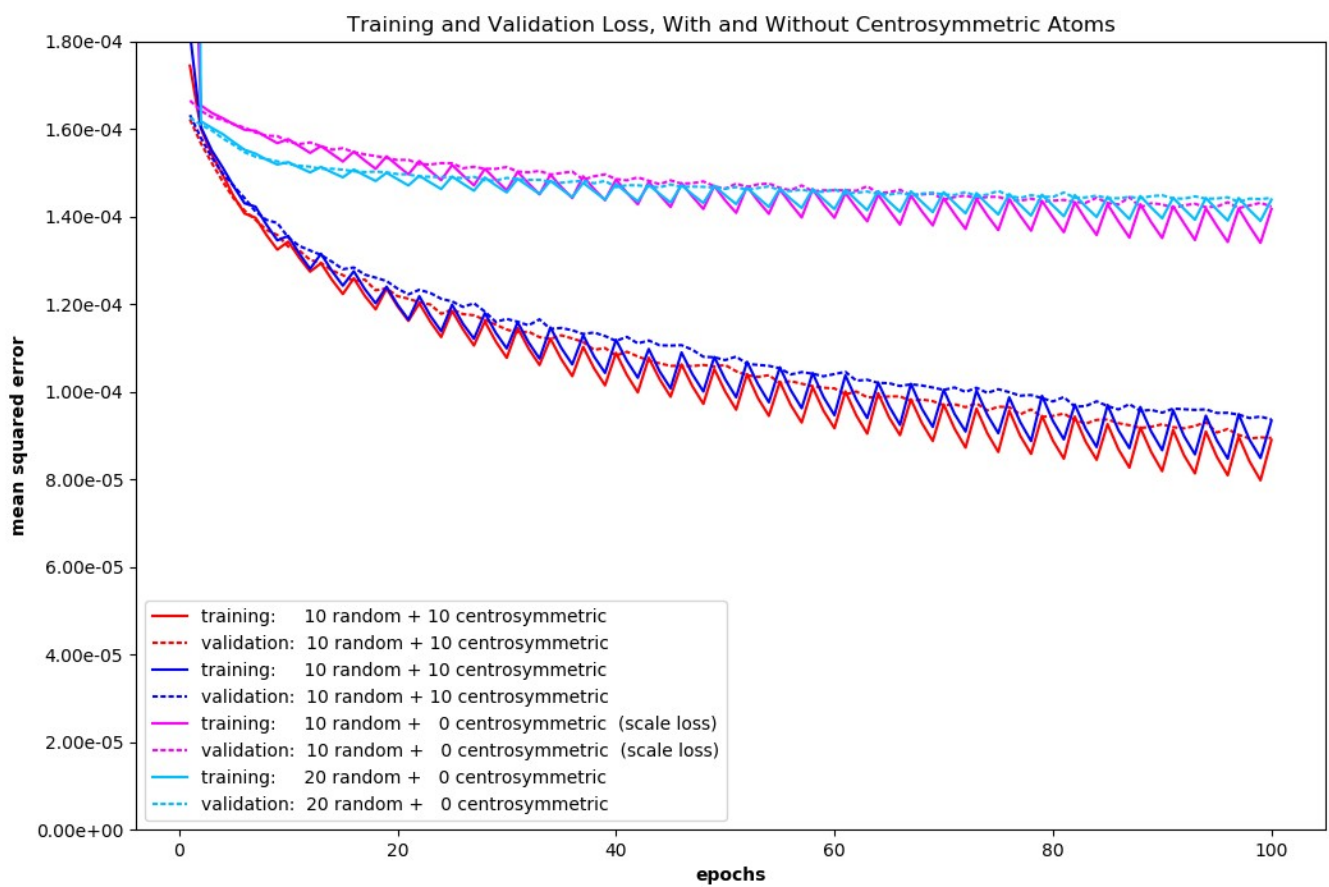

Fig. 8. This figure shows 100 epochs of training for a neural network trained in 2 ways. In the usual case, the network trained on both 10 random atoms and their centrosymmetric counterparts simultaneously. These are the darker lines, where the loss decreases much faster. When the network trained on just random atoms (no centrosymmetry atoms), training and generalization suffered large performance degradation. These are the lighter colored lines. In all cases, the training curves are zig-zagged because new training cases were swapped in every 3 epochs. 300 validation cases were created at the start of training and used throughout. These are the dashed lines.

## 3.2.2 Training Without Removing Ambiguity about Vector Origins

For the second test, the restriction that the random atom centers were confined to an inner box of size [12 x 12 x 12] pixels$^3$ inside a larger box of [40 x 40 x 40] pixels$^3$ was relaxed. These tests were performed to check the claim made in Methods 2.4.3 that empty space must be added around the atoms for the network to train and generalize. In this test, I trained 2 new neural networks from scratch, simply increasing the inner box size from [12 x 12 x 12] to either [18 x 18 x 18] or [24 x 24 x 24] pixels$^3$. Results for this test are shown in figure 9 below, demonstrating that the neural network does not train or generalize when the inner box is [24 x 24 x 24] pixels$^3$, and only partially when the inner box is [18 x 18 x 18] pixels$^3$. The same 2 baseline tests were used in figures 8 and 9. An explanation for the need to add empty space around the atoms is given in Methods 2.4.3.

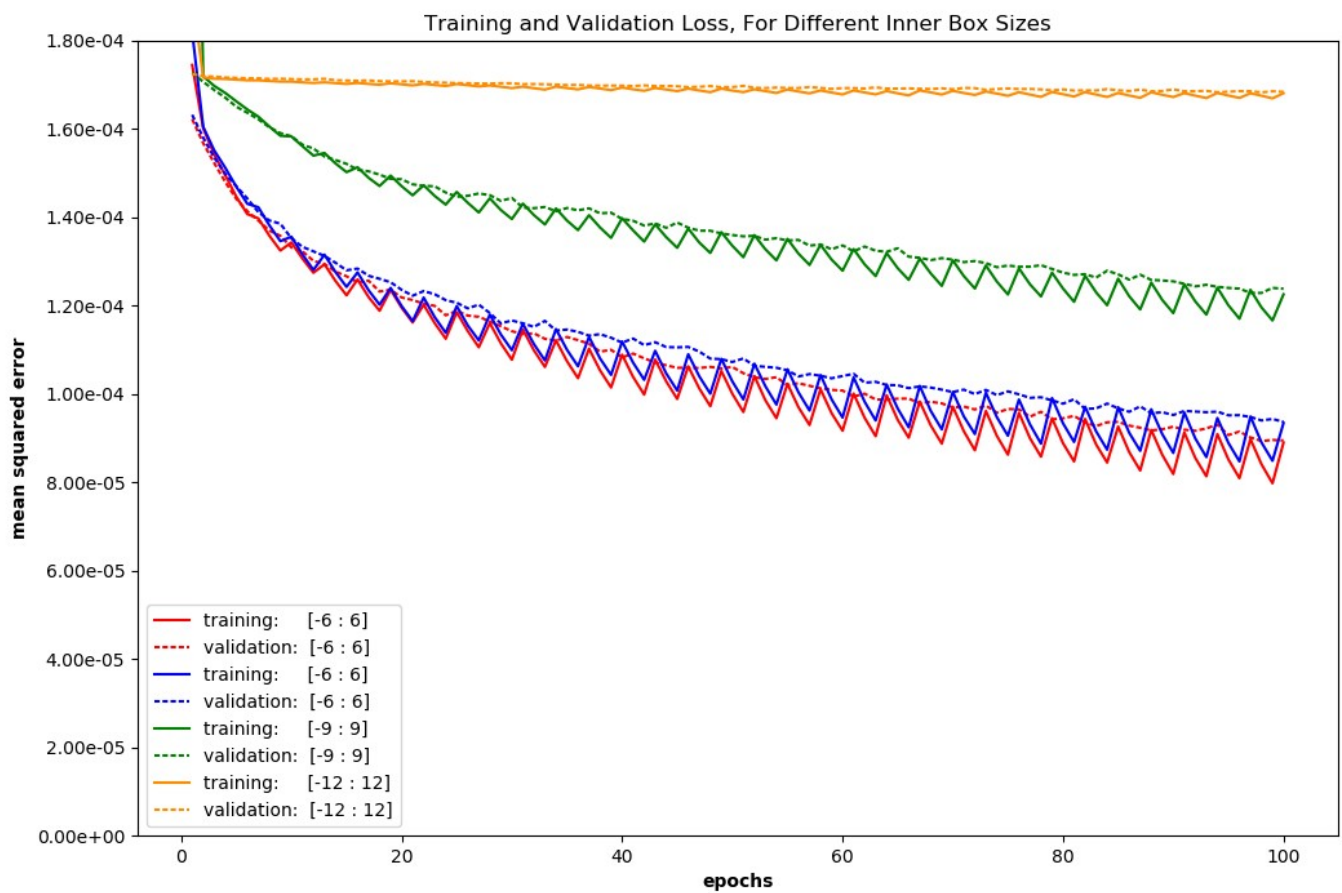

Fig. 9. This figure shows 100 epochs of training for 4 neural networks trained from scratch. The networks were trained using 10 randomly-positioned atoms plus their centrosymmetric counterparts. The inner box containing the 20 atoms was modified for these tests. In 2 cases, the usual inner box size of [12 x 12 x 12] was used. These are the red and blue colored lines where the loss decreases much faster. For the other 2 cases, the inner box size was increased to either [18 x 18 x 18] (green), or [24 x 24 x 24] (orange). As the inner box size increased, the network's ability to train and generalize decreased. In all cases, the training curves are zig-zagged because new training cases replaced the old ones every 3 epochs. 300 validation cases were created at the start of training and used throughout. These are the dashed lines.

## 3.3 Output From the Trained Network

During the inference stage, the trained neural network was presented with Patterson maps that it had not seen before. The Patterson maps were derived from synthetic density maps, as described in Methods 2.2. In figure 10, true synthetic density maps are visually compared to outputs from the trained neural network. In all cases that I have seen, the predicted outputs and the original density maps appear strikingly similar.

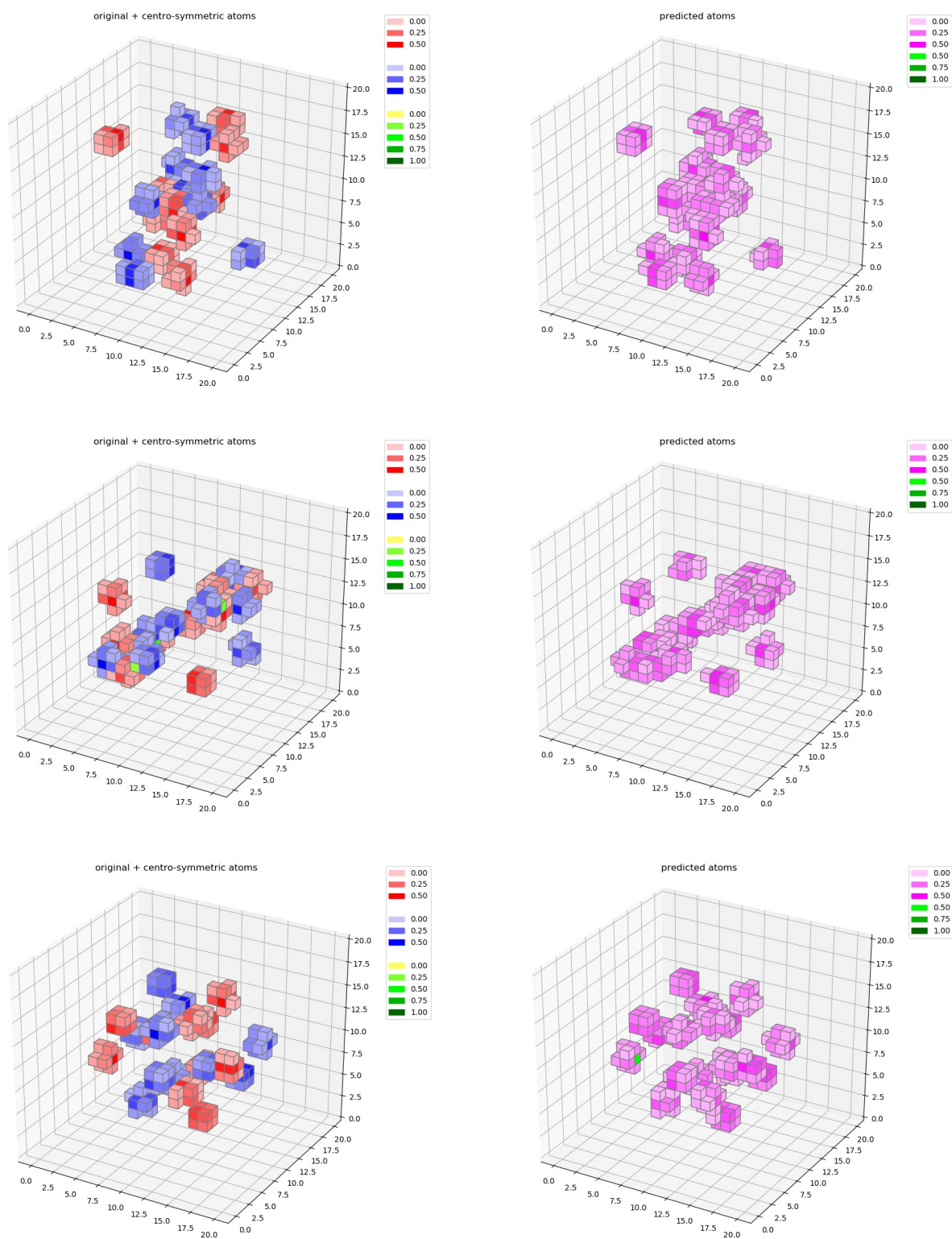

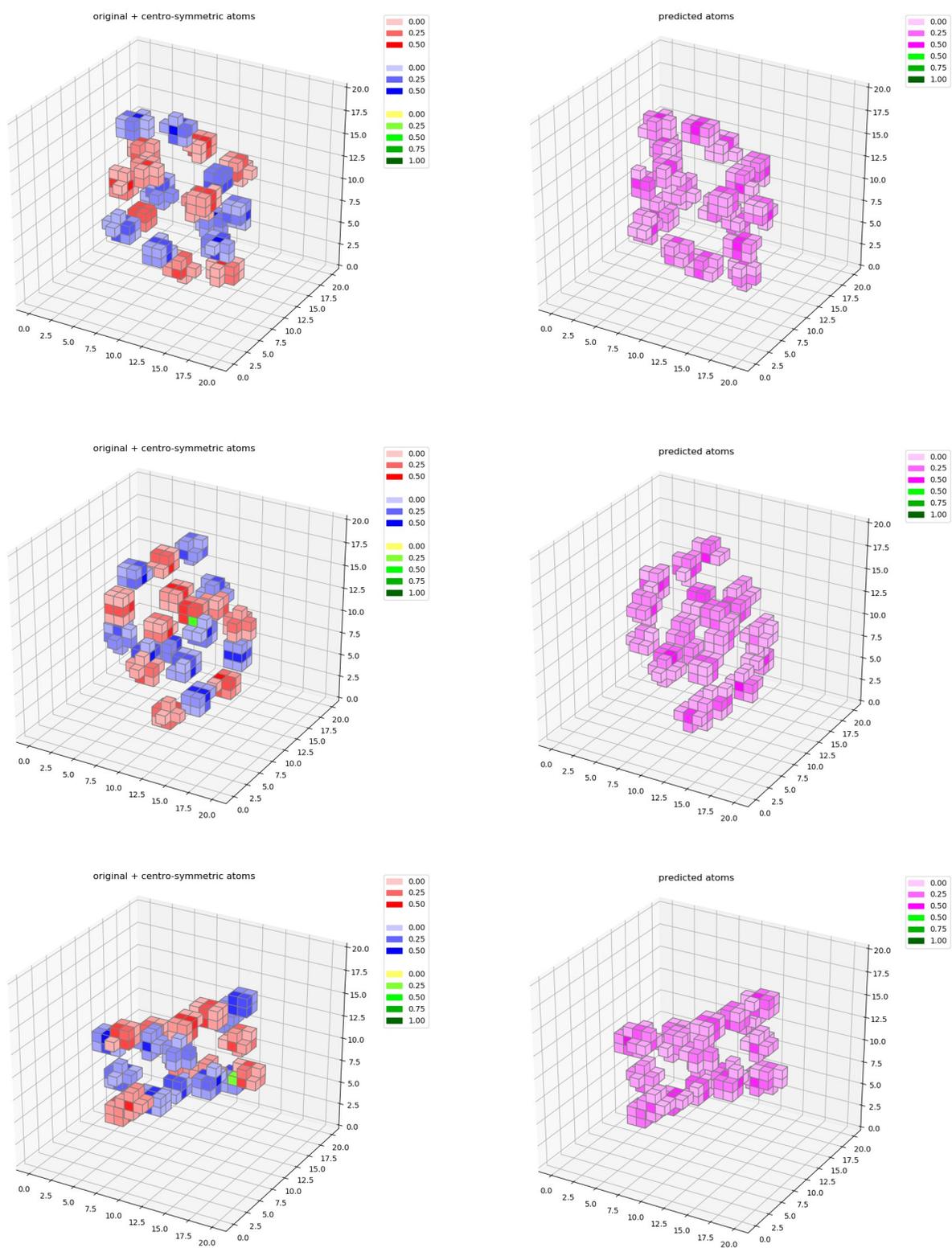

Fig. 10. In this figure, true outputs on the left are compared to inferred neural network outputs on the right. These examples are for cases that were not in the training set. The true outputs have 10 original atoms (red) plus 10 centrosymmetric atoms (blue). Clashes are allowed between the red and blue atoms (green). On the right are neural network outputs which are later interpreted to give atomic coordinates. For clarity, pixels with less than 10% occupancy are left empty.

## 3.4 Separating Two Sets of Atoms

This neural network solves for original atoms plus centrosymmetric atoms *simultaneously.* As a result, it is necessary to separate the atoms on the network output into 2 sets after running the inference stage. The algorithm to do this (Methods 2.4.4) initially chooses, at random, 10 of the positions that can be determined from the neural network output. The Patterson map for this set of 10 atoms is calculated and compared to the true Patterson map. The goal is to choose a set of 10 atoms whose Patterson map best matches the true Patterson map by swapping atoms, as described in Methods 2.4.4. Figure 11 shows the improvement of the mean-squared-error between test and true Patterson maps, for 7 test cases, as the algorithm that swaps atom positions runs and settles on a set of 10 atoms.

To be clear, this method cannot distinguish between a set of atoms and their centrosymmetric counterparts. This method can just as easily settle on the original atoms used to make the Patterson map as their centrosymmetrically-related atoms. This is because both sets of atoms have the same Patterson map.

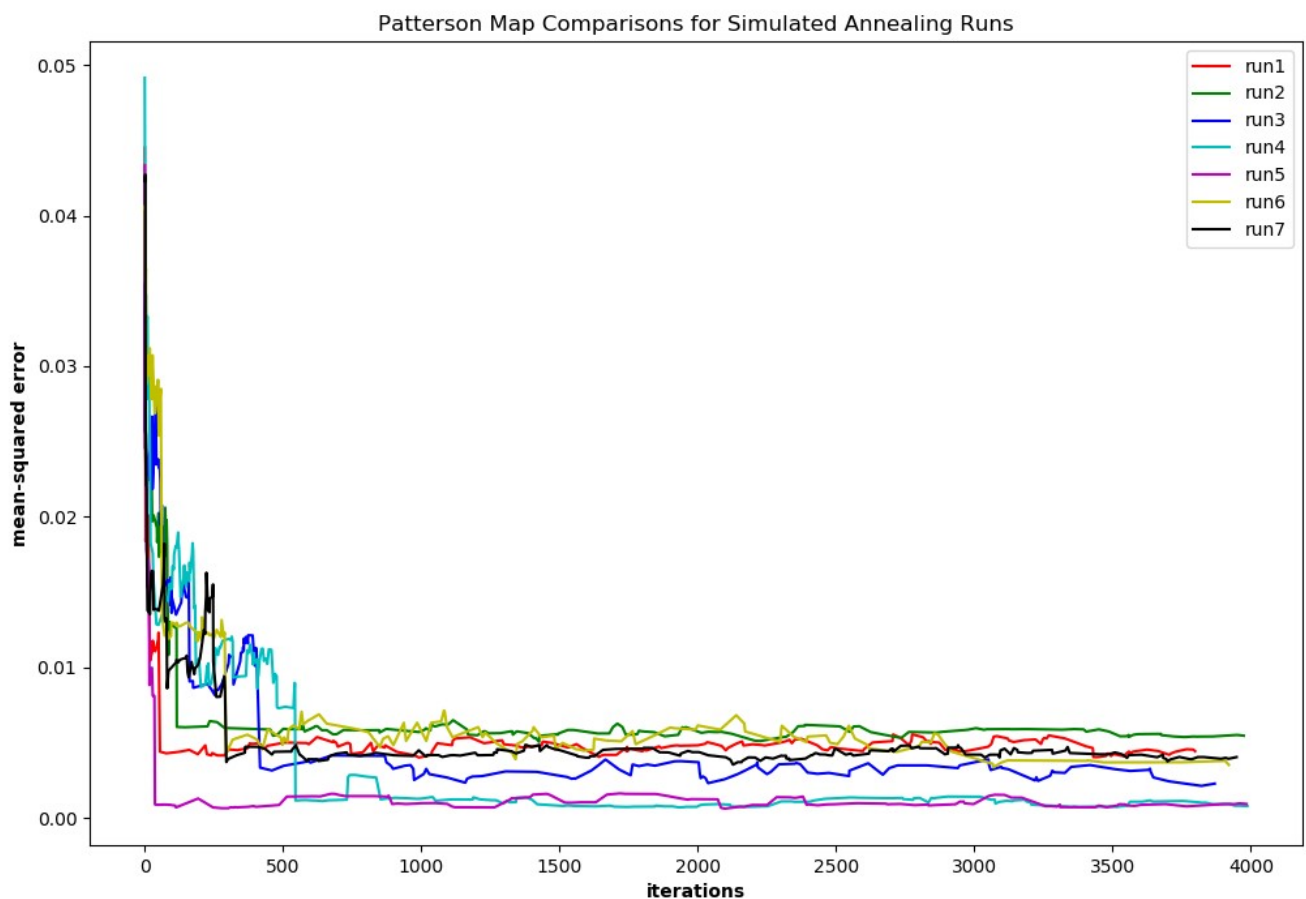

Fig. 11. This figure shows the mean-squared-error between true Patterson maps and Patterson maps derived from neural network outputs, for 7 test cases. Patterson maps from the network outputs are made from 10 of the 40 or so atoms positions determined from the network output maps. As the composition of atoms in the set of 10 improves, the similarity score between the Patterson maps improves as well.

## 3.5 Positional Accuracy

For testing, 500 true density maps were made. A Patterson map was calculated for each density map and presented to the neural network input. The trained neural network was tested by generating inferred density maps from the true Patterson maps and comparing them to the true density maps.

Atom positions were determined from the neural network outputs by taking weighted position-averages of a peak pixel and its neighbors. The separation algorithm described in Methods 2.4.4 was then used to determine 10 positions whose Patterson map best matched the true Patterson map.

This set of 10 atom positions was compared to the true atom positions to determine positional accuracy. Since the 10 atoms and their centrosymmetric counterparts produce the same Patterson map, it is not possible to tell which set of 10 deduced atoms to use when calculating positional accuracy. For this test, if positional accuracy was poor when comparing deduced atom positions to known atom positions, the signs on the deduced atom positions were flipped, thereby comparing the deduced atoms' centrosymmetry-related positions to the known atom positions instead. So this test, in essence, compares both deduced atoms and their centrosymmetric counterparts to the true atom positions and discards the poorer match of the two.

Deduced and true positions were only matched if a true position was the closest one to a deduced position, and vice-versa. In all, of the 5000 atom positions in the 500 density maps, 4994 were matched. Figure 12 shows a histogram of positional accuracy for the 500 test maps. Positional accuracy was, on average, 0.283 pixels for the 4994 matched atoms.

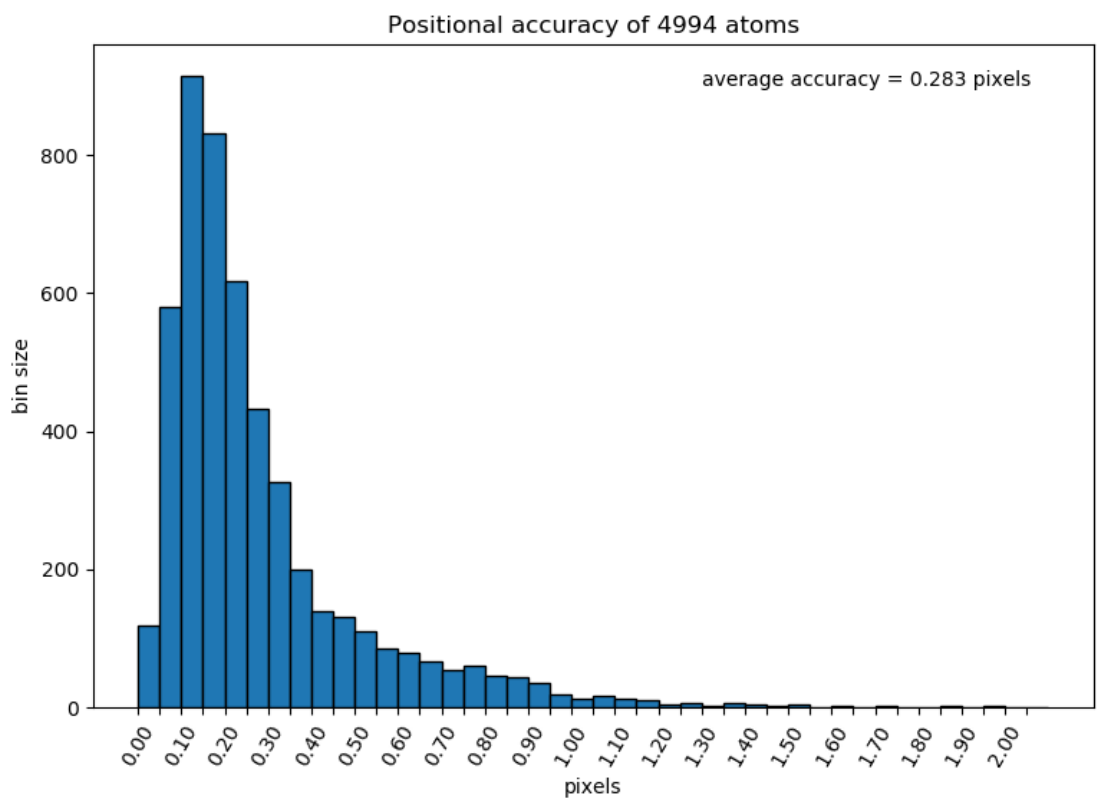

Fig. 12. This figure is a histogram of the positional accuracy of 4994 atoms deduced from 500 neural network output maps. For each output map, 10 positions were determined and compared to the 10 positions used to make the Patterson map. For 6 of the 5000 positions, a position comparison could not be made because the match between original and deduced atoms was unclear. Average accuracy is 0.283 pixels.

# 4. Discussion

In this work, a neural network was successfully trained to infer atom positions from Patterson maps. Looking at figure 10, the similarity between true and inferred density maps is striking. I compared many additional plots like these, and in all cases the output maps from the inference stage were qualitatively very similar to the true density maps that were used to create the Patterson map inputs. Additionally, positional accuracy of under 1 pixel for the large majority of atoms, with only a few missing atoms, is quite a good result as well (figure 12).

It occurs to me that, though each training and test case was made from randomly positioned atoms, perhaps this neural network is functioning as a large memory device. The network was trained on roughly 3 million training cases; the network could have a memory of all of those training cases and simply interpolate between them for the test cases. One test that can be done is to remake the network, reserving a portion of atomic configuration space for validation and testing. For example, when making the output density maps, I could impose the condition that, during training, a corner of this space must remain empty of atoms. Then, this restriction could be removed during validation and testing. This is something to keep in mind for future work. Only time prevented me from testing this for this paper, as retraining the network takes weeks.

In a sense, though, it's fine if the network functions as a large memory device, interpolating between cases stored in its memory during training, so long as it works during testing. The concern really is that, as configuration space grows exponentially, the network won't be able to remember a sufficient number of test cases, and the method will stop working. So, another way to test this concern will be to simply scale up the number of atoms and the space they reside in, and see if the network continues to generalize.

Aside from simply scaling up the number of atoms, eventually I would like to train and test this network on real data, to see if this method can infer atomic coordinates from experimental X-ray diffraction data. To do this, it will be necessary to train the network using electron density maps on the network output, and their Patterson maps on the network input.

Many open questions remain. This method may not scale up in size. In addition, even if it scales up in size, it may not work on protein structures whose atoms are not positioned randomly. Additionally, the strategy of adding empty space around the random atoms (Methods 2.4.3), which was necessary for this method to work, may not work with real data.

For all these reasons, the strategy of deducing atomic coordinates from Patterson maps that is outlined in this paper, and was used successfully here on a small-scale synthetic problem, may not be directly applicable to solving protein structures from X-ray data. Still, this work is a first step toward a neural network approach to solving the phase problem. Some of the insights from this work may be helpful in future work.